\pdfoutput=1
\documentclass[11pt]{article}

\usepackage[final]{acl}

\usepackage{times}
\usepackage{latexsym}
\usepackage{float}
\usepackage{tcolorbox}

\usepackage[T1]{fontenc}
\usepackage[utf8]{inputenc}
\usepackage{booktabs}
\usepackage{caption}

\usepackage{microtype}
\usepackage{colortbl} % Add this in your preamble for the color support
\usepackage{inconsolata}
\usepackage{array} % Add this in the preamble

\usepackage{graphicx}
\usepackage{tcolorbox}
\usepackage{amssymb} % For tick and cross symbols
\usepackage{pifont}
\newcommand{\xmark}{\ding{55}}%
\usepackage{makecell} % Add this to the preamble

\usepackage{xcolor}
\definecolor{darkgray}{rgb}{0.4, 0.4, 0.4}
\definecolor{backcolour}{rgb}{0.95,0.95,0.92}
\definecolor{myblue}{rgb}{0.2, 0.4, 0.8} % Example blue color
\definecolor{mygreen}{rgb}{0.2, 0.6, 0.2} % Example green color
\definecolor{myteal}{RGB}{0,128,128}
\definecolor{mylavender}{HTML}{E6E6FA}

\newcommand{\redcell}{\cellcolor{red!30}}

\title{\texttt{TemporalVQA}: Can Multimodal LLMs do Visual Temporal Understanding and Reasoning? The answer is No!}

\author{Mohamed Fazli Imam\textsuperscript{1}\quad Chenyang Lyu\textsuperscript{2}\quad Alham Fikri Aji\textsuperscript{1} \\
\textsuperscript{1}Mohamed bin Zayed University of Artificial Intelligence\\
\textsuperscript{2}Alibaba International Digital Commerce\\
\texttt{\{mohamed.imam,alham.fikri\}@mbzuai.ac.ae}\quad\texttt{lyuchenyang.dcu@gmail.com}
}

\begin{document}
\maketitle

\begin{abstract} 
Multimodal Large Language Models~(MLLMs) have achieved significant advancements in tasks like Visual Question Answering~(VQA) by leveraging foundational Large Language Models~(LLMs). However, their abilities in specific areas such as visual temporal understanding, which is crucial for comprehending real-world dynamics, remain underexplored. To address this, we propose a challenging evaluation benchmark named \texttt{TemporalVQA}, consisting of two parts: 1) \textbf{Temporal Order Understanding} and 2) \textbf{Time-lapse Estimation}. The first part requires MLLMs to determine the sequence of events by analyzing temporally consecutive video frames. The second part presents image pairs with varying time differences, framed as multiple-choice questions, asking MLLMs to estimate the time-lapse between images with options ranging from seconds to years. Our evaluations of advanced MLLMs, including models like GPT-4o and Gemini-1.5-Pro, reveal significant challenges: GPT-4o achieved only 49.1\% average consistent accuracy in temporal order task and 70\% in time-lapse estimation, with open-source models performing even poorly. These findings underscore the limitations of current MLLMs in visual temporal understanding and reasoning, highlighting the need for further improvements for their temporal capability. 
Our dataset can be found at \textcolor{red}{\url{https://huggingface.co/datasets/fazliimam/temporal-vqa}}.
\end{abstract}

\begin{figure}[!t]
    \centering
    {
        \includegraphics[width=\linewidth]{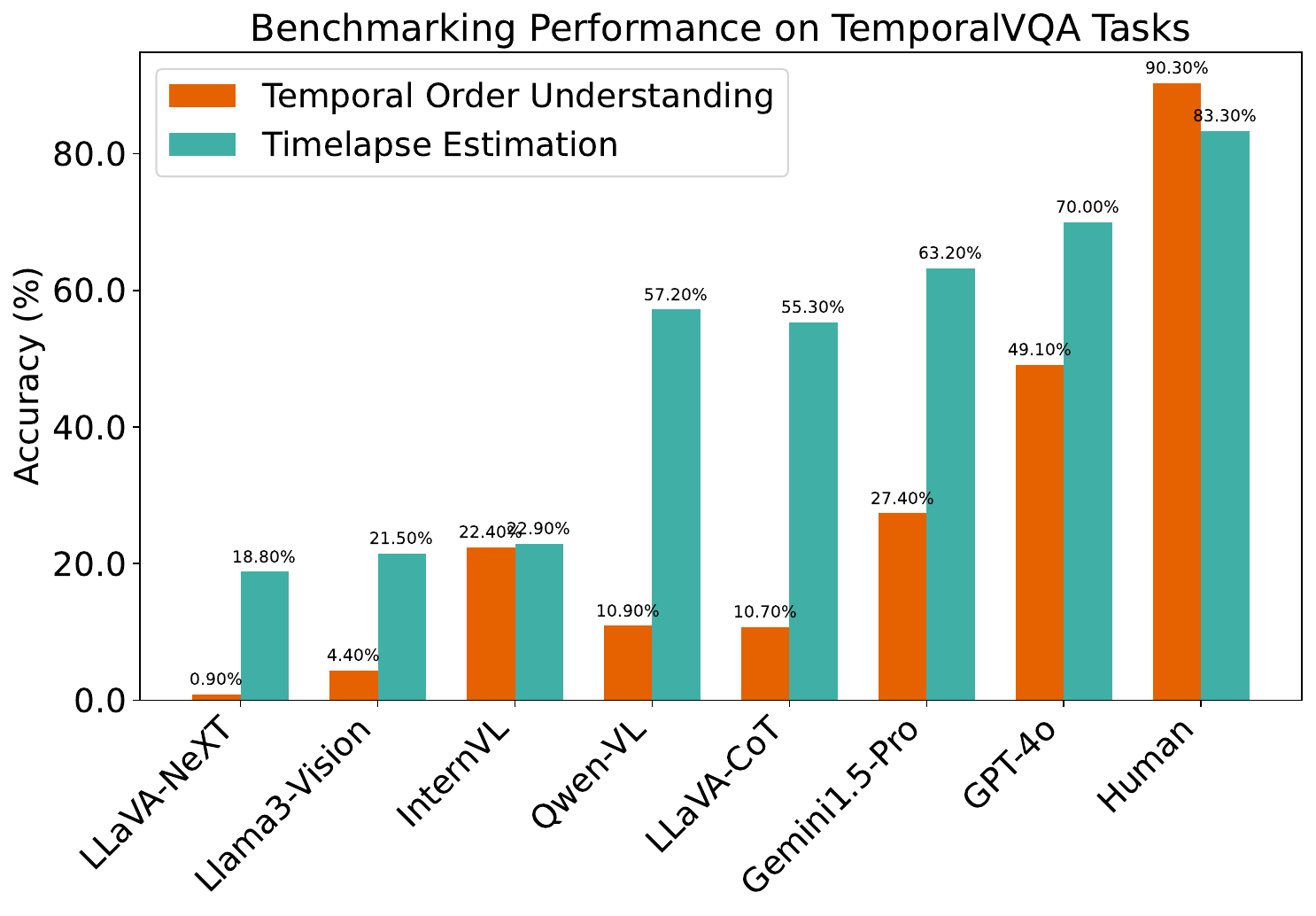}
    } 
    \vspace{-20pt}

    \caption{Performance comparison across TemporalVQA tasks. The plot shows the accuracy (\%) of different models on Temporal Order Understanding (\textcolor[HTML]{E66100}{orange}) and Timelapse Estimation (\textcolor[HTML]{40B0A6}{green}). The accuracy shown for Temporal Order Understanding is averaged consistent accuracy across all different layouts. 
    % Human performance is significantly higher across both tasks, while GPT-4o is the best-performing model among the MLLMs. Other models exhibit varying levels of success, with noticeable performance gaps between the two tasks.
    }
    \label{fig:perf_comp_plot}
\end{figure}

\section{Introduction}

Multimodal Large Language Models~(MLLMs) have achieved significant progress in various multimodal tasks, notably in Visual Question Answering (VQA), by leveraging the capabilities of foundational Large Language Models (LLMs)~\cite{yin2023survey_mllm,li2024surveybenchmarksmultimodallarge}. While these models are strong at understanding and interpreting visual scenes and information, their abilities in specific areas such as temporal understanding and reasoning have not been thoroughly explored~\cite{huang2024surveyevaluationmultimodallarge}. Temporal understanding is crucial because it evaluates how well MLLMs comprehend the motion and dynamics of real-world events~\cite{wallat2024_temporal_blind_spot_llms}. 
Without effectively understanding the sequence and timing of events, MLLMs miss out on grasping the dynamics of real-world scenarios, which are crucial for tasks that involve interpreting changes over time and predicting future states accurately. 

Consequently, enhancing temporal reasoning in MLLMs is essential to bridge this gap, enabling them to perform more effectively in applications such as video analysis, autonomous navigation, and time-sensitive decision-making. By advancing their temporal comprehension, MLLMs can achieve a more holistic understanding of the world, leading to more accurate predictions and interactions in dynamic environments.

To address this gap, we introduce a simple yet challenging evaluation benchmark named \texttt{TemporalVQA}, which consists of two parts: \textbf{Temporal Order Understanding} and \textbf{Time-lapse Estimation}.

Our experimental results reveal that \texttt{TemporalVQA} poses a significant challenge for current advanced MLLMs. For the \textbf{Temporal Order Understanding} task, GPT-4o~\cite{hurst2024gpt4o}, averaged across three layouts, achieved only 49.1\% average consistent accuracy, which is equivalent to random guessing. Open-source models like LLaVA-NeXT~\cite{liu2023llava} performed even worse with less than 10\% accuracy. 

For the \textbf{Time-lapse Estimation} task, advanced MLLMs such as GPT-4o and Gemini only achieved around 70\% and 63.2\% accuracy, respectively. As shown in Figure~\ref{fig:perf_comp_plot}, even the most advanced models, such as GPT-4o and Gemini-1.5-Pro, struggled to achieve satisfactory performance, with GPT-4o reaching only 49.1\% averaged consistent accuracy in temporal order understanding and 70.0\% in time-lapse estimation tasks, substantially below human performance of 90.3\% and 83.3\% respectively. These results underscore the limitations of current MLLMs, particularly in their ability to perform satisfactory temporal understanding and reasoning. 

In this paper, we aim to highlight the limitations of existing MLLMs in handling temporal understanding and reasoning tasks. We will discuss the presented benchmark, the challenges that MLLMs face in temporal tasks, and the need for more robust evaluation frameworks. Our study emphasizes the importance of developing advanced models and benchmarks to improve the temporal reasoning abilities of MLLMs.

\begin{figure*}[htbp]
    \centering
    \includegraphics[width=0.65\linewidth]{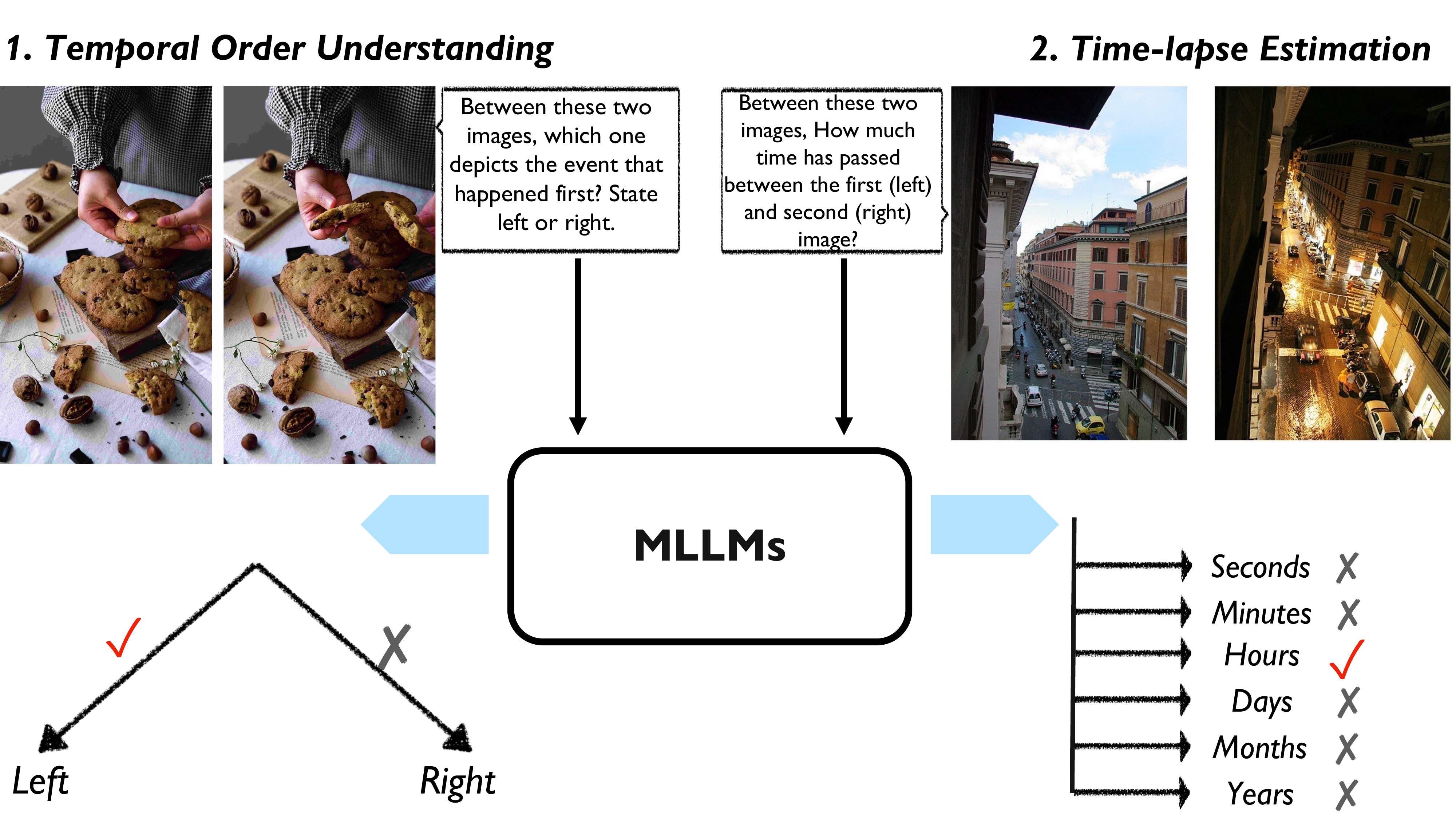}
\caption{An introductory diagram illustrating the task setup for the \texttt{TemporalVQA} benchmark. In \textbf{Temporal Order Understanding}, the model is asked to determine which of the two images depicts the event that happened first. In \textbf{Time-lapse Estimation}, the model estimates the time duration between two images selecting from options like seconds, minutes, hours, days, weeks/months or years.} \label{fig:temporal_bench_intro} 
\end{figure*}

\section{Related Works}
There have been works discussing the temporal understanding capability for text-based LLMs~\cite{chu-etal-2024-timebench,fatemi2024testtimebenchmarkevaluating,wang-zhao-2024-tram}, but there still lacks a thorough examination of the temporal capabilities for MLLMs.

The existing multimodal benchmarks~\cite{yue2023mmmu,cai2024temporalbench} predominantly focus on single-image inputs or consist of straightforward questions that do not challenge the models' deeper understanding or reasoning capabilities. Most benchmarks mainly examine the relationship between the visual and textual components, lacking a thorough exploration of the interactions among elements within the visual input~\cite{li2023seedbench,liu2024mibenchevaluatingmultimodallarge}. 

These limitations mean these benchmarks~\cite{fu2023mme,fu2024video_mme} do not adequately assess the temporal reasoning skills of MLLMs, which are essential for accurately understanding sequences of events and dynamic contexts. The ability to discern the order and timing of events is vital for applications ranging from temporal visual analysis to real-time decision-making~\cite{sun2024temporalinsightenhancementmitigating}.

Recently, two benchmarks, TOMATO~\cite{shangguan2024tomatoassessingvisualtemporal} and TVBench~\cite{cores2025tvbenchredesigningvideolanguageevaluation}, have been introduced. TVBench aims to refine video-language evaluation, advocating for better-designed benchmarks that mitigate bias and reliance on static cues.
TOMATO evaluates MLLMs on video understanding through Multi-Frame Gain, Frame Order Sensitivity, and Frame Information Disparity, focusing on sequential video frame analysis. While TOMATO  examines how models interpret video frames as a continuous sequence, our benchmark, \textbf{TemporalVQA} focuses on the ability to determine event sequences and estimate time differences between images.

\section{TemporalVQA Benchmark}

We construct the \texttt{TemporalVQA} benchmark to evaluate the temporal understanding capabilities of multimodal large language models~(MLLMs). The benchmark consists of two main tasks: \textbf{Temporal Order Understanding} and \textbf{Time-lapse Estimation}, as illustrated in Figure~\ref{fig:temporal_bench_intro}

\subsection{Task Formulation}

\paragraph{Temporal Order Understanding}
In this task, we provide the model with two images, where one precedes the other in terms of event order. The model is then asked to determine which image comes first. To ensure robustness, we present the same image pair twice, swapping their order each time, effectively doubling the data size.

\paragraph{Time-lapse Estimation}
In this task, we provide the model with two images depicting an ongoing event (e.g., people cooking), where the first image precedes the second. The model is then asked to estimate the time elapsed between the two images. We define six distinct labels, ranging from mere seconds to events spanning several years.

\subsection{Image Source}

\paragraph{Temporal Order Understanding} For the Temporal Order Understanding task, we sourced videos from copyright-free platforms, including Pixabay~\footnote{\url{https://pixabay.com/}} and Pexels~\footnote{\url{https://www.pexels.com/}}. From these videos, we sampled pairs of frames that depict a clear sequence of events, ensuring enough distinction between the frames to make the temporal order evident. For example, one pair consists of a frame showing an empty cup, followed by a frame with the cup filled with tea, clearly illustrating the chronological sequence. The statistics for this task are presented in Table~\ref{tab:temp_order_data_stats}.

\begin{table}[h]
\centering
\small
\renewcommand{\arraystretch}{1.2}
\resizebox{0.86\linewidth}{!}{
\begin{tabular}{ll}
\toprule
\textbf{Dataset Attribute} & \textbf{Value} \\
\midrule
Total Image Pairs         & 720 \\
Unique Image Pairs        & 360 \\
Label Options             & \textit{2 (First / Second)} \\
Categories                & \redcell \makecell[l]{\textit{Human Activities, Sports, Cooking} \\ \textit{Baking, Animal Motion, Plant Growth} \\ \textit{Object Transformations}} \\
\bottomrule
\end{tabular}
}
\caption{Dataset statistics for the Temporal Order Understanding task.}
\label{tab:temp_order_data_stats}
\end{table}

\paragraph{Time-lapse Estimation}
For the Time-lapse Estimation task, we curated image pairs by selecting copyright-free images and videos from platforms such as Flickr~\footnote{\url{https://flickr.com/}}, Google~\footnote{\url{https://images.google.com/}}, and YouTube~\footnote{\url{https://youtube.com/}}. Each pair features scenes or objects captured at varying time intervals, ranging from seconds and minutes to years. The images were carefully chosen to provide clear visual cues that indicate the time elapsed between them. Examples include images of a fruit before and after ripening, or a landscape scene across different seasons. 
The statistics for this task are presented in Table~\ref{tab:timelapse_data_stats}.

\begin{table}[h]
\centering
\scriptsize
\renewcommand{\arraystretch}{1.2}
\resizebox{0.86\linewidth}{!}{
\begin{tabular}{ll}
\toprule
\textbf{Dataset Attribute} & \textbf{Value} \\
\midrule
Pairs                   & \textit{125} \\
Labels                  & \textit{6} \\
Examples per Label       & \cellcolor{blue!30} \makecell[l]{\textit{Seconds - 19, Minutes - 20, Hours - 20,} \\ \textit{Days - 18, Weeks/Months - 19, Years - 29}} \\
Categories              & \redcell \makecell[l]{\textit{Human Activities, Sports, Cooking \& Baking,} \\ \textit{Animal Motion, Plant Growth,} \\ \textit{Object Transformations, Environmental Changes}} \\
\bottomrule
\end{tabular}
}
\caption{Dataset statistics for the Timelapse estimation task.}
\label{tab:timelapse_data_stats}
\end{table}

\subsection{Data Annotation}

To ensure high-quality data, we follow up with manual filtering and annotations. For Temporal Order Understanding, filtering is done to ensure there are no ambiguous entities. For Time-lapse Estimation, annotation is performed to set the label.

\paragraph{Temporal Order Understanding} Several image pairs were sampled from the sourced videos, each sampled at 1 fps. From the sampled frames, two or more frames were selected to ensure a clear distinction between the two events, avoiding ambiguous transitions and ensuring one event clearly preceded the other. Depending on the nature of the video, the time gap between the selected frames varied: for slow-motion videos, the frames were spaced further apart, while for fast-paced videos, the frames were closer together.

Each image pair was reviewed by two independent annotators. Annotators manually filtered pairs that represented a clear event sequence, with the first frame labeled as the \textbf{earlier event} and the second as the \textbf{later event}. If visual cues (e.g., object movement, lighting changes) were insufficient to establish order, the pair was discarded. Frames with watermarks, timestamps, or text indicating time were excluded, and to prevent reliance on superficial cues, frames with excessive scene similarity were avoided unless motion-based progression was evident.

\paragraph{Time-lapse Estimation} Similarly, for the time-lapse estimation task, both videos and images illustrating "before" and "after" scenarios were sourced to create the dataset. Images clearly showing these distinctions were paired accordingly. For the videos, the process followed the same methodology as the temporal order understanding task. 

After generating the image pairs, each pair was labeled by two annotators, who classified them into six predefined time ranges. Annotators were provided with metadata, such as the video source, title, and description, which sometimes offered clear labels for the time-lapse categories. In cases where the metadata alone was insufficient, annotators used visual indicators such as lighting conditions, object movement, seasonal changes, and biological transformations to decide on the appropriate time-lapse category. 
If annotators encountered difficulty assigning a category due to insufficient temporal cues, the pair was flagged for further review. In cases of disagreement or if an image pair was too ambiguous to categorize, the pair was resolved through discussion or discarded, with preference given to pairs that exhibited clear temporal logic and objective visual cues.

\subsection{Prompt Design}

\label{sec:prompt_design}
\paragraph{Temporal Order Understanding}
To evaluate the consistency of the models, we generated multiple dataset variations, with image pairs arranged in horizontal and vertical layouts. Additionally, within each version, we created sets by swapping the positions of the images to assess the models' sensitivity to spatial arrangement. Based on these variations, we designed two types of prompts. In one prompt, the task is framed as a true or false question:

\begin{tcolorbox}[title=Prompt \#1]
\textit{Did the event in the \textcolor{red}{(\textbf{left} / \textbf{top} / \textbf{first})} image happen before the event in the \textcolor{red}{(\textbf{right} / \textbf{bottom} / \textbf{second})} image? State \textbf{true} or \textbf{false} with reasoning.}
\end{tcolorbox}
In the other prompt, the task is to specify which image depicts the event that happened first:
\begin{tcolorbox}[title=Prompt \#2]
\textit{Between these two images, which one depicts the event that happened first? State \textcolor{red}{(\textbf{left} or \textbf{right} / \textbf{top} or \textbf{bottom} / \textbf{first} or \textbf{second})} with reasoning.}
\end{tcolorbox}

\paragraph{Time-lapse Estimation}
For this task, we framed the questions in a multiple-choice format, asking MLLMs to estimate the time-lapse between the two images. The options range from seconds, minutes, hours, days, months, and years. To minimize ambiguity in cases where closely related timeframes could both be plausible, we rephrased the options to introduce a sufficient gap between them. 

Unlike the previous task, which required two different prompts to assess event chronology across varying layouts, a single prompt is sufficient for this task. Since the task focuses on estimating the duration between events rather than determining their relative order, there is no need to evaluate consistency across different layouts. Instead, the model's performance is assessed solely based on its ability to select the most appropriate time interval from the given options, making a single, well-structured prompt sufficient for accurate evaluation:

\begin{tcolorbox}[title=Prompt \#3]
\textit{In the given image, estimate the time that has passed between the first image (left) and the second image (right). Choose one of the following options: \\
\textbf{\textcolor{red}
        {A. Less than 15 seconds\\
        B. Between 2 minutes to 15 minutes\\
        C. Between 1 hour to 12 hours\\
        D. Between 2 days to 30 days\\ 
        E. Between 4 months to 12 months\\
        F. More than 3 years}}}\\
        \textit{Provide your answer with reasoning.}
\end{tcolorbox}

\section{Evaluation Results}

\begin{table*}[h!]
\centering
\resizebox{\textwidth}{!}{
\begin{tabular}{lcccccccccccccc}
\toprule
 & \multicolumn{2}{c}{\textbf{GPT-4o}} & \multicolumn{2}{c}{\textbf{Gemini1.5-Pro}} & \multicolumn{2}{c}{\textbf{LLaVA-NeXT}} & \multicolumn{2}{c}{\textbf{InternVL}} & \multicolumn{2}{c}{\textbf{Qwen-VL}} & \multicolumn{2}{c}{\textbf{Llama3-Vision}} & \multicolumn{2}{c}{\textbf{LLaVA-CoT}}  \\
 & P1 & P2 & P1 & P2 & P1 & P2 & P1 & P2 & P1 & P2 & P1 & P2 & P1 & P2\\ \midrule
\multicolumn{13}{l}{\textit{Vertically Joined}} \\ 
Correct answer on the top (1st order) & 93.9\% & 94.7\% & 30.8\% & 81.4\% & 25.3\% & 100\% & 71.4\% & 85.3\% & 6.1\% & 44.4\% & 6.4\% & 5.25\% & 84.2\% & 91.4\%\\ 
Correct answer on the bottom (2nd order) & 32.2\% & 40.6\% & 88.3\% & 41.9\% & 71.9\% & 0.0\% & 27.2\% & 17.8\% & 96.9\% & 66.4\% & 93.6\% & 46.1\% & 16.7\% & 13.3\%\\ 
Average Accuracy & 63.1\% & 67.6\% & 59.6\% & 61.7\% & 48.6\% & 50.0\% & 49.3 \% & 51.5\% & 51.5\% & 55.4\% & 50.0\% & 49.3\% & 50.4\% & 52.4\%\\
\rowcolor{lightgray} Consistent Accuracy & 31.4\% & 39.2\% & 23.6\% & 35.0\% & 6.4\% & 0.0\% & 16.4\% & 13.6\% & 5.8\% & 19.2\% & 3.3\% & 7.8\%  & 13.1\% & 12.8\%\\ \hline
\multicolumn{13}{l}{\textit{Horizontally Joined}} \\ 
Correct answer on the left (1st order) & 95.0\% & 92.8\% & 52.5\% & 70.8\% & 7.5\% & 100\% & 66.4\% & 79.4\% & 10.8\% & 100\% & 3.3\% & 33.3\% & 83.6\% & 89.7\%\\ 
Correct answer on the right (2nd order) & 29.4\% & 49.4\% & 62.8\% & 44.4\% & 89.2\% & 0.0\% & 39.4\% & 22.8\% & 92.8\% & 0.0\% & 95.6\% & 64.2\% & 18.6\% & 12.5\%\\ 
Average Accuracy & 62.2\% & 71.1\% & 57.6\% & 57.6\% & 48.3\% & 50.0\% & 52.9\% & 51.1\% & 51.8\% & 50.0\% & 49.4\% & 48.8\% & 51.1\% & 51.1\%\\
\rowcolor{lightgray} Consistent Accuracy & 30.6\% & 52.8\% & 21.4\% & 29.4\% & 4.2\% & 0.0\% & 25.6\% & 18.6\% & 8.3\% & 0.0\% & 1.9\% & 4.4\% & 14.2\% & 10.3\%\\
\midrule
\multicolumn{13}{l}{\textit{Separate Multiple Images}} \\
Correct answer is the first image (1st order)& 81.9\% & 85.6\% & - & - & 0.3\% & 100\% & 72.5\% & 67.2\% & 0.6\% & 50.8\% & - & - & 35.0\% & 94.2\%\\
Correct answer is the second image (2nd order)& 54.2\% & 70.8\% & - & - & 99.4\% & 0.0\% & 28.6\% & 40.8\% & 99.4\% & 59.4\% & - & - & 69.2\% & 6.7\%\\
Average Accuracy & 68.1\% & 78.2\% & - & - & 49.9\% & 50.0\% & 50.6\% & 54.0\% & 50.0\% & 55.1\% &  - & - & 52.1\% & 50.4\%\\
\rowcolor{lightgray} Consistent Accuracy & 43.6\% & 65.3\% & - & - & 0.0\% & 0.0\% & 22.2\% & 26.4\% & 0.6\% & 23.9\% &  - & - & 14.7\% & 4.7\%\\

\bottomrule
\end{tabular}}

\caption{Temporal order understanding results across different models and prompting strategies. Models are evaluated on joined image pairs (vertically and horizontally) and seperated image pairs (for applicable models). For each layout, we report the accuracy of temporal ordering in both directions (Top-to-Bottom/Bottom-to-Top for vertical; Left-to-Right/Right-to-Left for horizontal) and the consistent accuracy (when model predictions are consistent across both directions). P1 and P2 represent two different prompting strategies as described in Section~\ref{sec:prompt_design}.}
\label{tab:temporal_order_results}
\end{table*}

\subsection{Evaluated MLLMs}
In our experiments, we evaluated a series of MLLMs, including strong closed-source models such as GPT-4o~\cite{hurst2024gpt4o} and Gemini-1.5-Pro~\cite{geminiteam2023gemini}, as well as state-of-the-art open-source MLLMs like LLaVA-NeXT~\cite{liu2023improvedllava}, InternVL~\cite{chen2024internvlscalingvisionfoundation}, Qwen-VL~\cite{bai2023qwenvlversatilevisionlanguagemodel}, Llama-3-Vision~\cite{dubey2024llama3herdmodels}, and LLaVA-CoT~\cite{xu2024llavacotletvisionlanguage}.

\subsection{Experimental Setup}

We conducted a series of experiments on our \texttt{TemporalVQA} benchmark to evaluate the temporal reasoning capabilities of several state-of-the-art MLLMs. 

For the \textbf{Temporal Order Understanding task}, each image pair was combined either vertically or horizontally to form a single input, challenging the models to determine which frame represents the earlier event and which one the later event. 

Since this task was framed as a binary question, with random performance potentially reaching 50\%, we introduced a "consistent accuracy" metric to ensure a more reliable assessment of the models' true temporal reasoning abilities. This metric evaluates whether a model provides the same prediction when the image pairs are presented in reversed order. If a model's predictions vary between the presented order, the response is marked as incorrect, preventing inflated accuracy due to random guessing. 

For a prediction to be considered both correct and consistent, the model must provide identical answers for both orders in the specified layout. This approach ensures that models are not only making correct predictions but also demonstrating robustness in their temporal understanding across different visual presentations.

On the other hand, in \textbf{Time-lapse Estimation task}, the models are required to estimate the approximate time duration between the image pairs, offering a more fine-grained measure of their temporal reasoning capabilities. This task provided deeper insights into whether the models could not only recognize temporal sequences but also approximate the passage of time between events.

The decoding strategies vary based on model architectures. GPT-4o uses an API call with a \texttt{max\_tokens} limit of 512 for multimodal inputs, while Gemini 1.5 Pro and InternVL employ structured pipelines with iterative decoding, also capped at 512 tokens. LLaMA and LLaVA tokenize inputs and use beam search, whereas LLaVA-COT extends this with reasoning-aware decoding, extracting both reasoning steps and conclusions. QWEN-VL utilizes an image-conditioned chat template with autoregressive decoding, refining outputs by trimming unnecessary tokens. 

For GPT-4o, Gemini 1.5 Pro, InternVL, Qwen-VL, and LLaVA-CoT, we extract predictions directly from the response. For LLaVA-NeXT and LLaMA-3-vision models, we treat the task as a multiple-choice question and select the option (A or B) with the highest likelihood based on next token probability, following the evaluation method in the original MMLU~\cite{hendryckstest2021_mmlu}. 

Across all models, deterministic decoding with \texttt{num\_beams=1} and controlled token generation is used to ensure consistency and reliability in responses.
\begin{table*}[!t]
\centering
\resizebox{0.9\linewidth}{!}{
\begin{tabular}{lcccccccc}
\toprule
\textbf{Time Span} & \textbf{Human} & \textbf{GPT-4o} & \textbf{Gemini1.5-Pro} & \textbf{LLaVA-NeXT} & \textbf{Qwen-VL} & \textbf{InternVL} & \textbf{Llama3-Vision} & \textbf{LLaVA-CoT}\\ 
\midrule
\quad Seconds & 87.7\% & 89.5\% & 78.9\% & 78.9\% & 63.2\% & 26.3\% & 100\% & 94.7\%\\ 
\quad Minutes & 78.9\% & 80.0\% & 90.0\% & 10.0\% & 65.0\% & 75.0\% & 5.0\% & 10.0\% \\ 
\quad Hours & 76.5\% & 40.0\% & 60.0\% & 0.0\% & 30.0\% & 20.0\% & 0.0\% & 30.0\% \\ 
\quad Days & 85.2\% & 61.1\% & 44.4\% & 0.0\% & 77.8\% & 11.1\% & 0.0\% & 88.9\%\\ 
\quad Weeks/Months & 85.4\% & 63.2\% & 47.4\% & 0.0\% & 21.1\% & 5.3\% & 0.0\% & 63.2\%\\ 
\quad Years & 86.2\% & 86.2\% & 58.6\% & 24.1\% & 86.2\% & 0.0\% & 24.1\% & 44.8\%\\ 
\midrule
\textbf{Average Acc.} & 83.3\% & 70.0\% & 63.2\% & 18.8\% & 57.2\% & 22.9\% & 21.5\% & 55.3\% \\ 
\bottomrule
\end{tabular}
}
\caption{Time-lapse estimation accuracy across different temporal scales. Models are evaluated on their ability to estimate time intervals between image pairs, ranging from seconds to years. The task requires models to select the appropriate time scale that best describes the temporal gap between two given images. }
\label{tab:timelapse_results}
\end{table*}

\subsection{Task-1: Temporal Order Understanding Results}

 Overall, the experimental results shown in Table~\ref{tab:temporal_order_results} reveal that all current MLLMs, including GPT-4o, the most advanced model in our evaluation, struggle with the proposed benchmark. Despite GPT-4o's superior performance relative to other models, it fails to consistently demonstrate accurate temporal reasoning across different settings. The consistent accuracy scores are notably low for all models, indicating significant limitations in their ability to comprehend and interpret temporal sequences from visual inputs. 
 
 These deficiencies are evident even when models are provided with multi-image inputs or optimized prompts, suggesting that current architectures and training methodologies are insufficient for robust temporal order understanding. 

\paragraph{Effect of different prompts} 
Our experiments show significant performance variations across prompts. GPT-4o exhibits some sensitivity, improving from P1 to P2 (31.0\% to 46.0\% in single-image, 43.6\% to 65.3\% in multi-image settings). 

However, even with optimized prompts, its performance remains close to random guessing. In contrast, models like LLaVA-NeXT and Qwen-VL show high sensitivity, with performance dropping under alternate prompts. This suggests that while prompt engineering influences outcomes, it cannot fully overcome limitations in temporal reasoning.

\paragraph{Effect of the layout of images within single-image input}
The image layout (vertical vs. horizontal) substantially affects model performance, exposing their limitations. GPT-4o shows better consistency in vertical layouts, improving from 39.2\% to 52.8\% with P2, though P1 sees a negligible drop (31.4\% to 30.6\%). Most models exhibit strong directional bias—LLaVA, for instance, performs perfectly in some orientations but fails completely when reversed. This inconsistency suggests models rely on spatial cues rather than truly understanding temporal sequences. 

\paragraph{Effect of multi-image input and single-image input}
The comparison between multi-image and single-image approaches reveals interesting patterns. GPT-4o shows slight improvement with multi-image input (from 31.0\% to 43.6\% using P1 and from 46.0\% to 65.3\% using P2), yet the overall performance remains limited. Other models exhibit mixed results, with InternVL maintaining relatively stable but low performance across both settings, and Qwen-VL showing minor improvements in P2 with multi-image input (9.6\% to 23.9\%). 

These findings suggest that simply providing additional visual context through multiple images does not significantly enhance temporal reasoning capabilities. The models appear to struggle with integrating temporal information, regardless of the input format.

\paragraph{Comparison with human performance}
For Task‑1, we conducted a human evaluation by engaging three independent participants to answer the questions. In this survey, participants were presented all 360 image-pairs with Prompt-2~(P2) as the question, and they were asked to select either the left or right image. 

Additionally, to avoid positional bias, the survey alternated the presentation order of image pairs. Specifically, 50\% of the pairs were displayed in the \textbf{\textit{Correct answer on the left (1st order)}}, while the remaining 50\% were in the \textbf{\textit{Correct answer on the right (2nd order)}}. 

The comparison between the best-performing MLLM for Task-1~(in our experiments, GPT-4o) and human performance of average participants accuracy is shown in Table~\ref{tab:bestmodelvhuman}. Even the best-performing system is still under human performance by \textbf{25\%}.

Upon analyzing the survey results, there was no single case where \textbf{all} participants made an incorrect prediction while there were individual misclassifications. That demonstrates that the dataset is reliable and consistent, with human participants largely agreeing on the correct predictions.

\begin{table}[!t]
\centering
\small % Reduce font size
\begin{tabular}{lc}
\toprule
Best Model (GPT-4o) & 65.3\% \\
\rowcolor{lightgray} Avg Human Evaluation & 90.3\% \\
\bottomrule
\end{tabular}
\caption{Accuracy comparison of the best performing MLLM with human performance.}
\label{tab:bestmodelvhuman}
\end{table}

\begin{figure*}[!t]
    \centering
    \includegraphics[width=0.95\linewidth]{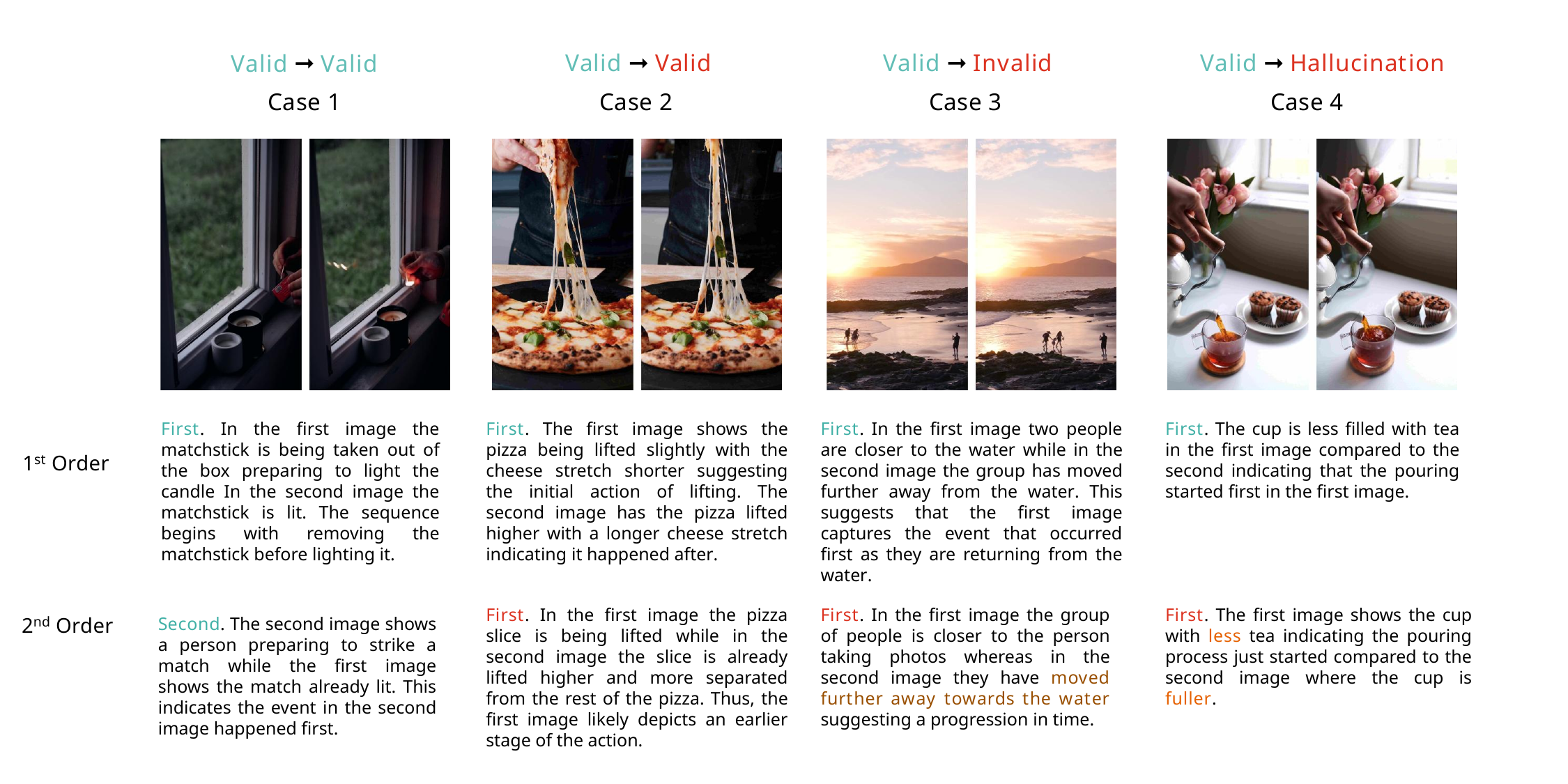}
    \vspace{-15pt}
    \caption{Some qualitative cases illustrating the output predictions from GPT4o. \textbf{1st order} refers to cases where the image pairs are fed to the model in their original sequence, while \textbf{2nd order} refers to cases where the image pairs are fed in reverse (swapped) order. Text highlighted in \textcolor[HTML]{40B0A6}{green} represents correct classifications while \textcolor[HTML]{DC3220}{red} indicates misclassifications. \textcolor[HTML]{E66100}{Orange} denotes instances of hallucinations, and \textcolor[HTML]{994F00}{brown} denotes instances of illogical reasoning.} 
    \label{fig:qual_results} 
\end{figure*}

\subsection{Task-2: Time-lapse Estimation Results}
As shown in Table~\ref{tab:timelapse_results}, all MLLMs, including top performers like GPT-4o and Gemini1.5-Pro, struggle with this task, achieving only moderate accuracy levels of 60-70\%. GPT-4o shows inconsistent performance, with strong performance in \textit{Seconds} and \textit{Years} but underperforming in \textit{Hours}. 

Similarly, LLaVA-CoT demonstrates exceptional performance in the time spans of \textit{Seconds} and \textit{Days}, while showing notably poor performance in the other time intervals. 
LLaVA-NeXT achieved competitive performance in estimating short intervals (\textit{Seconds}) but struggled with longer time spans, indicating a reliance on fine visual details over temporal reasoning. Qwen-VL's high accuracy in \textit{Years} contrasts with its poor performance in shorter intervals, revealing a focus on significant temporal changes rather than fine-grained understanding. Other models, like InternVL and Llama3-Vision, exhibit low overall accuracy, with isolated strengths like Llama3-Vision's performance in \textit{Seconds}. 

The human evaluation for Task-2 followed the same procedure as Task-1. All 125 image pairs were used in the survey for three participants, and the evaluation questions were formed using Prompt-3~(P-3). As observed from the results in Table, the average human performance surpassed the best system - GPT-4o, by 13.3\%. 

A closer inspection of the survey responses shows that there was only one instance where all participants provided an incorrect answer. This reaffirms that the image pairs are clearly defined and unambiguous as well as the overall reliability and quality of the evaluation data for Task-2.

\subsection{Case Study}
Among all models and setups, GPT-4o with \textbf{P2} in the multi-image configuration performed best. Hence, this configuration was chosen to assess the soundness of the reasoning behind its predictions. 

\begin{table}[h!]
    \centering
    \resizebox{0.9\linewidth}{!}{
    \begin{tabular}{lc}
        \toprule
        \textbf{Consistency Rate} & \textbf{74.2\% (267)} \\
        \hspace{1em}\textit{Consistent correctly classified} & \textcolor{blue}{\textit{65.3\% (235)}} \\
        \hspace{1em}\textit{Consistently misclassified} & \textcolor{red}{\textit{8.9\% (32)}} \\
        \bottomrule
    \end{tabular}
    }
        \caption{Breakdown of GPT-4o's consistency rate: correctly classified (\textcolor{blue}{blue}) and incorrectly classified (\textcolor{red}{red}), with sample counts in brackets.}
\label{tab:gpt4o_analysis_1}
\end{table}

As shown in Table \ref{tab:gpt4o_analysis_1}, the model achieves a consistently correct classification rate of 74.2\%, the ability of the model to alter its predictions appropriately when the images are presented in a swapped manner, with the majority (65.3\%) attributed to its ability to correctly classify consistently. 

Similarly, the model has a consistently incorrect classification rate of 8.9\%, which was due to misidentifying objects and elements in the image pairs.

To verify that consistent accuracy is a valid and reliable measure of the model's performance, we sampled 30 instances from the 267 consistently correctly classified cases and evaluated the model's reasoning when presented with image pairs in both layouts. In all 30 instances, the reasoning was deemed valid for both orders. This outcome aligns with the expectation that the model exhibits reasonably robust reasoning capabilities regardless of the order of the images. An example of this is illustrated in Figure \ref{fig:qual_results} under Case 1. 

We analyzed the reasoning behind the model’s inconsistent predictions, which occurred in 93 image pairs. Among these, 73 were correctly classified when the first image was the correct answer (1st order), while 20 were correctly classified only when the second image was the correct answer (2nd order). This highlights the model's sensitivity to image order when making predictions. 

Additionally, we examined the model's reasoning behind these inconsistencies. Specifically, we sampled 30 instances from the 73 pairs where the correct answer was the first image and analyzed all 20 pairs where the correct answer was the second image. A detailed breakdown of this analysis is presented in Table \ref{tab:gpt4o_analysis_2}. The reasoning provided by the model was categorized into three groups:

\begin{itemize} 
% \vspace{-1em}
    \item Valid - The provided reasoning is completely valid to determine the order of occurrence.
\vspace{-1em}
    \item Invalid - The provided reasoning is invalid or illogical to determine the order of occurrence.
\vspace{-2em}
    \item Hallucination - The provided reasoning includes imagined details about the image.
\end{itemize}

\begin{table}[!t]
\centering
\small
\resizebox{0.96\linewidth}{!}
{
\begin{tabular}{@{}lcc@{}}
\toprule
\textbf{Reasoning} & \textbf{1st \checkmark, 2nd \xmark} (73) & \textbf{2nd \checkmark, 1st \xmark (20)}\\
\midrule
Valid $\rightarrow$ Valid & 10.0\% (3) & 31.6\% (6) \\ 
Valid $\rightarrow$ Invalid & 66.7\% (20) & 15.8\% (3) \\ 
Valid $\rightarrow$ Hallucination & 20.0\% (6) & 2.0\% (1) \\ 
Invalid $\rightarrow$ Valid & 0 & 31.6\% (6) \\
Invalid $\rightarrow$ Invalid & 3.3\% (1) & 20.0\% (4) \\
Invalid $\rightarrow$ Hallucination & 0 & 0 \\

\bottomrule
\end{tabular}}
\caption{Breakdown of changes in reasoning (valid or invalid) when the order of the image pairs fed to the model is swapped.}
\vspace{-1em}
\label{tab:gpt4o_analysis_2}
\end{table}

Our analysis reveals that the model provides valid reasoning for the majority of cases (66.7\%) when the image pairs are presented in the 1st order. However, when the image pairs are presented in the 2nd order, the reasoning is evenly split between valid and invalid, highlighting the model's sensitivity to the input order.

As shown in Figure \ref{fig:qual_results} Case 2 (2nd Order), the model sometimes provides valid reasoning but with an incorrect prediction. This shows that the model's decision-making is influenced by the input order, leading to inconsistencies in its predictions. 

Alternatively, some predictions were incorrect and based on invalid reasoning. In Figure \ref{fig:qual_results} Case 3, a group of people is shown near the beach in the first image and closer to the photographer in the second, clearly indicating movement from the beach toward the photographer. When the image pair is presented in the correct order, the model provides valid reasoning and makes the correct prediction.
However, when the images are swapped and presented in reverse order, the model incorrectly reasons that the group is moving towards the water, demonstrating flawed reasoning influenced by the reversed input order.

Additionally, the model exhibited instances of hallucination in its predictions. In Figure \ref{fig:qual_results}, Case 4, when the image pair is presented to the model in the 2nd order, the model erroneously claims that the cup in the first image contains less tea than the second, even though it should be the other way around—the first image shows the cup with more tea. This highlights the model's tendency to generate inaccurate and imaginative reasoning in certain scenarios.

\section{Conclusion}
In this work, we introduced TemporalVQA, a novel benchmark designed to evaluate the temporal understanding and reasoning capabilities of MLLMs. Our comprehensive evaluation across multiple state-of-the-art MLLMs revealed significant limitations in their ability to perform visual temporal reasoning tasks. Our analysis exposed several critical challenges in current MLLMs: (1) high sensitivity to input order and layout, indicating a lack of robust temporal understanding; (2) inconsistent reasoning patterns, where models often provide valid reasoning but arrive at incorrect conclusions; and (3) tendency to derive illogical reasoning and hallucinate when making temporal judgments. These findings suggest that current MLLMs primarily rely on superficial visual cues rather than genuine temporal comprehension. Future work should focus on improving MLLMs to better capture temporal relationships and maintain consistency across different input configurations.

\section*{Limitations}
While our proposed \texttt{TemporalVQA} benchmark provides a valuable resource for evaluating the temporal understanding and reasoning capabilities of MLLMs, it still has some limitations: 1) the size of the dataset is relatively small compared to other large-scale benchmarks in the field. This may limit the generalizability of the findings and to fully reflect the ability of models of temporal reasoning. Expanding the dataset to include a larger and more diverse set of examples could provide a more comprehensive evaluation of MLLMs' temporal capabilities. 2) our benchmark primarily focuses on two specific tasks: temporal order understanding and time-lapse estimation. While these tasks are essential and crucial for temporal reasoning, more different task formats are needed to capture the full scope of temporal understanding required in real-world applications. Future work could explore additional temporal reasoning tasks, such as event duration estimation, temporal causality, and reasoning over long-term temporal dependencies. 3) our evaluations are limited to a subset of state-of-the-art MLLMs, including GPT-4o and Gemini-1.5-Pro. While these models represent the current state-of-the-art systems, some more sophisticated MLLMS can also be tested.

\bibliography{custom}

% \appendix

% \section{Appendix}
% \label{sec:appendix}

% This is an appendix.

\end{document}